\documentclass[10pt,twocolumn,letterpaper]{proc}

\usepackage{times}
\usepackage{epsfig}
\usepackage{graphicx}
\usepackage{amsmath}
\usepackage{amssymb}

\usepackage[detect-family]{siunitx}
\usepackage{relsize}
\usepackage{mathtools}
\usepackage{sansmath}
\usepackage{enumitem}
\usepackage{rotating}

\usepackage{lipsum}
\usepackage{blindtext}
\usepackage{comment}
\usepackage[dvipsnames]{xcolor}

\usepackage{algorithm}
\usepackage[noend]{algpseudocode}
\usepackage{multirow}

\begin{document}
\title{Image Denoising via CNNs: An Adversarial Approach}

\author{Nithish Divakar~~~~~~~~~~~R. Venkatesh Babu\\
Video Analytics Lab,\\
Dept. Computational and Data Sciences\\
Indian Institute of Science, Bangalore, India\\
}

\maketitle
\begin{abstract}
Is it possible to recover an image from its noisy version using convolutional neural networks?
This is an interesting problem as convolutional layers are generally used as feature detectors for tasks like classification, segmentation and object detection.
We present a new CNN architecture for blind image denoising which synergically  combines three architecture components,
a multi-scale feature extraction layer which helps in reducing the effect of noise on feature maps, an $\ell_p$ regularizer which helps in selecting only the appropriate feature maps for the task of reconstruction, and finally a three step training approach which leverages adversarial training to give the final performance boost to the model.
The proposed model shows competitive denoising performance when compared to the state-of-the-art approaches.
\end{abstract}

\section{Introduction}\label{sec:intro}

Image denoising is a fundamental image processing problem whose objective is to remove the noise while preserving the original image structure. 
%
%
Traditional denoising algorithms are given some information about the noise, but the problem of blind image denoising involves computing the denoised image from the noisy one without any knowledge of the noise.


Convolutional Neural Networks(CNNs) have generally been used for classification.
They have a set of convolutional layers(convolution followed by a non-linear function) and eventually a few fully connected layers which help in predicting the class.

But these networks have also found multiple other uses as the output of these convolutional layers provide a rich set of features from a seemingly nominal image.xBut what if these features are not exactly from an actual image, but something very close?
Can we reconstruct the clean image from features extracted from a noisy image?

This paper addresses how CNNs can be used for blind image denoising.
The problem does not fit into traditional frameworks as described above since input to the network is not clean images.

\begin{figure}[thbp]
\begin{center}
	\includegraphics[width=0.45\textwidth]{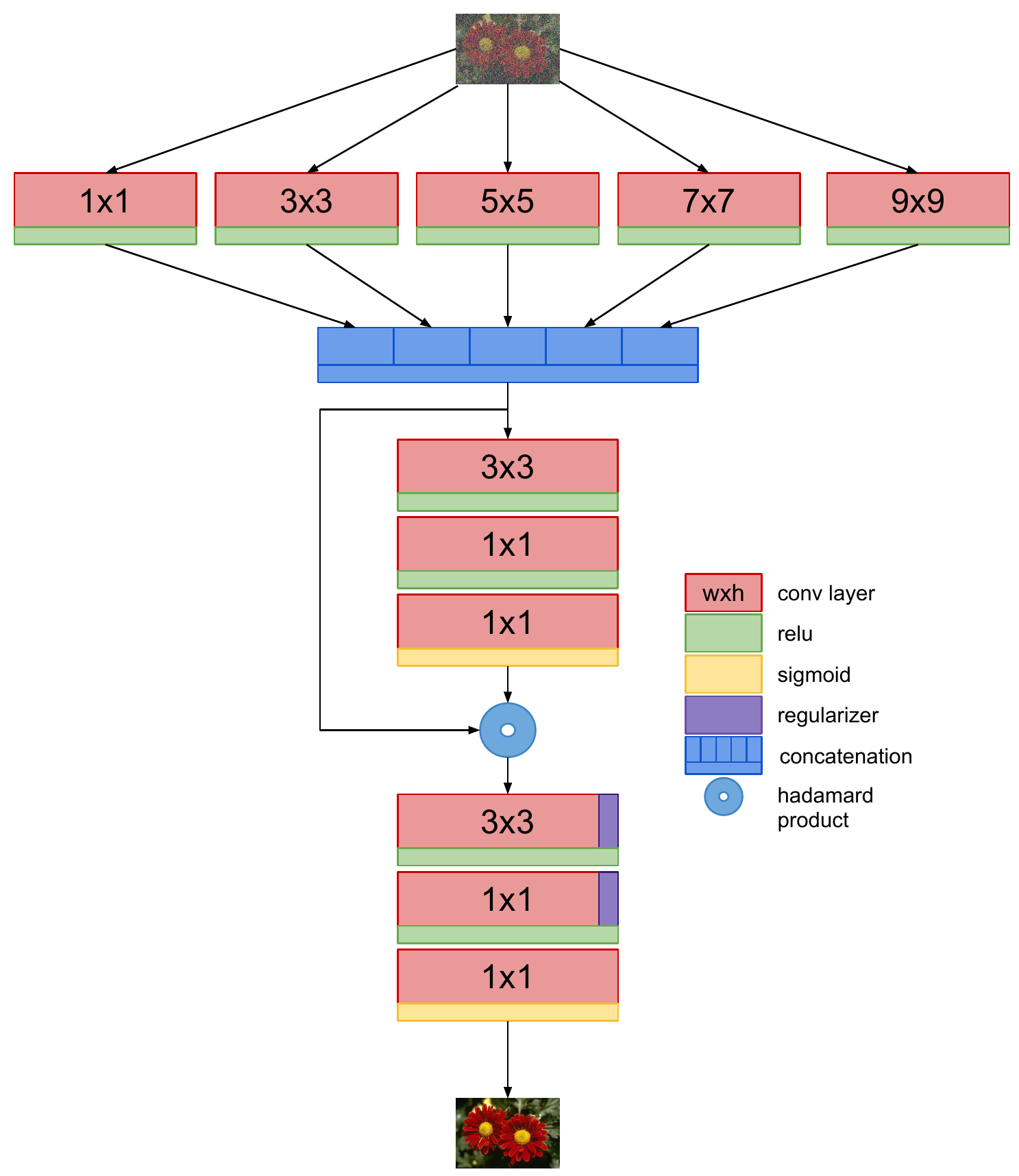}
\caption{The proposed image denoising model}
\label{fig:arch}
\end{center}
\end{figure}
They are noisy images and require the network to gather enough features from this image so that a noise-free version can be computed from them.
The proposed architecture is shown in Fig.~\ref{fig:arch}. It includes three main components 
(i) a set of filters that simultaneously extracts features at multiple scales from the image. We call these filters collectively as \textit{multi-scale feature extraction layer} 
(ii) a combination of filters which allow dampening the features contaminated by noise and 
(iii) reconstruction layers with filters that do not have any spatial resolution. The architecture is explained in detail in Sec.~\ref{sec:arch}.

	
The following are the major contributions of this paper.
\begin{itemize}
    \item We propose a multi-scale adaptive CNN architecture which gives a competitive performance to the state-of-the-art image denoising approaches.
    \item A training regime which exploits clean images as well as noisy images to get good feature maps for reconstruction.
    \item An adversarial training procedure, which helps to improve the denoiser performance further than the $\ell_2$ loss would allow. 
\end{itemize}


\section{Proposed approach}\label{sec:proposed}

The proposed denoising approach contains two main components: (i) an image denoising model  and (ii) a three phase training procedure. In this section, we present a detailed overview of both.

\subsection{Architecture of the denoiser}
\label{sec:arch}


Convolutional layers are traditionally used as feature detectors for the classification task. But stacking multiple convolutional layers on top of each other gives the network an inherent feature of abstracting details in deeper layers~\cite{gatys2015neural}. This property, although quite useful for classification and other related tasks, is unsuitable for image denoising as the finer details of the image need to be preserved for a good reconstruction.

A naive solution might be to simply use deconvolutions~\cite{deconv-eccv-2014}. But this, in turn, imposes more burden on the network to learn to reconstruct details from an abstract representation of the image. Moreover, such a network requires a large number of layers and hence is harder to train.

To circumvent this, we use two techniques.
\begin{enumerate}
    \item Extract as many features as possible from the image in the first layer itself.
    \item Keep all filters of the deeper layers to be $1 \times 1$ in size to avoid abstraction and blurring of fine image structures.
\end{enumerate}

To extract all of the necessary features from the image, simply having large number filters of the same size is not enough. Inspired from inception layers of GoogLeNet~\cite{Szegedy_2015_CVPR}, we employ multiple sets of convolutional filters, each set progressively having larger filter sizes, directly applied on the image. The resulting activations from all these layers are simply stacked together. We call the combination of these filters \textit{multi-scale feature extraction} layer.
\begin{figure}[ht]
    \centering
    \includegraphics[width=0.45\textwidth]{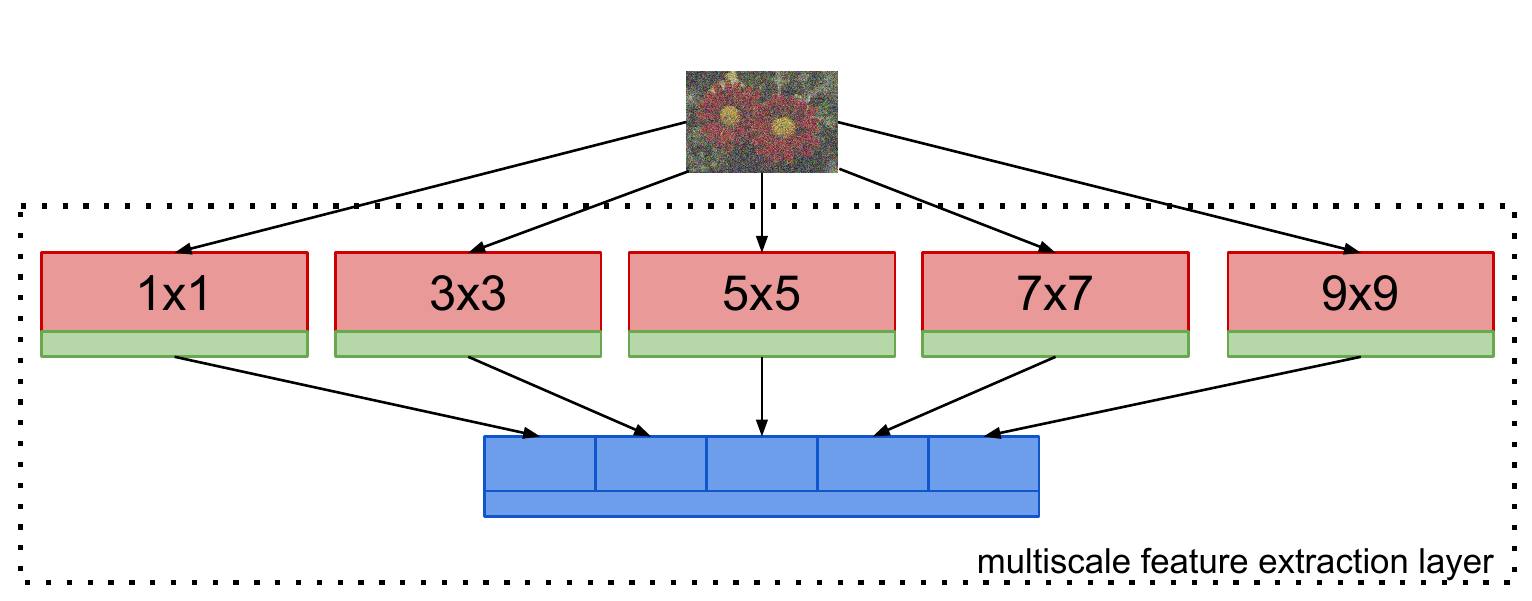}
    \caption{Multi-scale feature extraction layer}
    \label{fig:multifeat}
\end{figure}

The main difference of this layer from inception layer is the absence of initial $1 \times 1$ convolutions. Inception layers are usually fed activation of previous layers and hence receive multiple feature maps. In our case, these convolutional layers operate directly over the input image and hence do not require the initial $1 \times 1$ convolutions. We can say multi-scale feature extraction layer is more similar to naive inception layer \cite{Szegedy_2015_CVPR}.

Another difference is the number of output channels. Unlike inception layers which have the same number of output channels for each parallel paths, the conv-layers of multi-scale feature extraction layer has a progressive number of output channels since larger filter sizes can extract more information.
For our experiments, we have fixed the output channels to 32,40,48,56  and  64.

We avoid learning abstract features in the later layers of our model by limiting expressivity. To achieve this we limit the filter sizes of convolution layers to $1\times1$.

This results in our model having less number of parameters and also avoids blurring the fine image structures. Hence we are able to use a larger training dataset as opposed to many of the earlier works like \cite{Vemulapalli_2016_CVPR,roth2005fields}.
    
\subsection{Three phase training}\label{sec:training}

Simply training the model by feeding noisy images and constraining the output to be close to the clean image can cause the network to quickly converge to averaging out noise. To circumvent this, we make use of the clean images by first teaching the model to simply reconstruct from clean images and then to reconstruct from the noisy image. 
The training process involves the following:
\begin{enumerate}
   \item \textbf{Clean-to-clean reconstruction} Feed clean images to the model and train it to reconstruct the same image back.
   \item \textbf{Noisy-to-clean reconstruction} Feed noisy images to the model and train it to reconstruct the corresponding clean image back.
   \item \textbf{Adversarial training} Train the denoiser model using an adversarial strategy to increase the denoising performance.
\end{enumerate}

\textbf{Clean-to-clean reconstruction}

In the first phase of training, we leverage the availability of clean images to learn useful filters for image reconstruction.

The model is trained to reconstruct the clean image from itself. The intent of this phase is to allow the model to learn good features to reconstruct images. But to prevent the model from simply collapsing to an identity function, we apply a heavy dropout (p = 0.7) immediately after the multi-scale feature extraction layer. 

The middle three layers of the architecture in Fig.~\ref{fig:arch} are provided to dampen the activations of the first layer. The intuition is explained in the next training phase. Since the intent of this phase is to learn features for reconstruction, the skip connection over the middle three layers is short-circuited, resulting in these layers not being part of training. Essentially, we train a model of effective depth of 4 in this phase.

\textbf{Noisy-to-clean reconstruction} The next stage is training the network to reconstruct clean images from noisy images. The dropout added in the previous training phase is removed and the parameters of the \textit{multi-scale feature extraction} layer are frozen. But now, since the images are noisy, the quality of extracted feature maps is adversely affected for those learnt filters which are most sensitive to noise. Feature maps of those filters, which are invariant to noise remains the same. To aid in quick adjustment to these good and bad feature maps (in the context of denoising), we provide a few extra layers that allow to selectively reduce the effect of bad feature maps of the \textit{multi-scale feature extraction} layer.

These layers eventually output a value between 0 and 1 for each pixel position, when fed the activations of the \textit{multi-scale feature extraction} layer. These values are then point-wise multiplied (Hadamard multiplication between tensors) back to the feature maps. The features of the  \textit{multi-scale feature extraction} layer as result gets rescaled according to the value. A value close to 0 completely diminishes the feature map while a value of 1 simply allows it to pass unmodified. All the layers of this stage have 240 output channels.

We also impose an $l_p$ regularizer on the $5^{th}$ and $6^{th}$ layers (See Fig.~\ref{fig:arch}). These layers have filters of $1 \times 1$ and hence imposing a sparsity preserving regularizer will lead to the model selecting only a few connections between the layers. This is an automated way of selecting only a few good activation maps to reconstruct the image. The same idea was implemented in ~\cite{NIPS2008_3506} by only allowing a randomly chosen $8$ connections to the previous layer. We have found the value of $p=0.1$ to be satisfactory. Too low a value results in exploding loss function and too high a value results in model simply collapsing to pure averaging. Layers of this stage have $128$ output channels except for the last layer which has only 1.

\begin{figure}[htbp]
    \centering
    \includegraphics[width=0.45\textwidth]{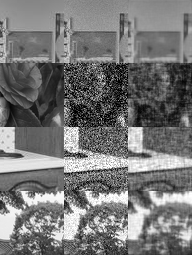}
    \caption{Denoising result after phase 2. The columns respectively show clean images, noisy images and reconstructed images.}
    \label{fig:mse-examples}
\end{figure}

Table~\ref{tab:mse-training} shows the denoising performances of the model at the end of this training phase. As can be inferred, the denoising performance is adequate, but far from the state-of-the-art results. Fig.~\ref{fig:mse-examples} shows some examples of denoised images at this stage of training for various noise levels. 

\begin{table}[htbp]
\caption{Denoising results after the end of \textit{noisy-to-clean} training phase on test set.}
    \centering
    \begin{tabular}{|lllll|}
\hline
Sigma     & 10    & 15    & 20    & 25\\\hline
PSNR      & 32.37 & 30.68 & 29.37 & 27.93 \\\hline
\end{tabular}
    \label{tab:mse-training}
\end{table}

We can see that the model just resorts to averaging all the pixel values in presence of heavy texture and high noise level. This effect can be attributed to the averaging effect of $\ell_2$ loss. For a detailed discussion, refer to ~\cite{mathieu2015deep}. To circumvent this effect, we need a better loss function that preserves natural image features. 

\textbf{Adversarial training}
Adversarial training of neural networks was introduced by  Goodfellow~et~al. in ~\cite{goodfellow2014generative}. We briefly describe it here.

Adversarial training is a method to train a generative network $G$ to generate samples from some real data $x\sim p_{data}$. Generators are fed input noise variables $z$ having distribution $p_Z$ and they are trained to learn the mapping to the data space. The distribution of the generator model is given by
\begin{align}
    p_g \sim G(z;\theta_g)
\end{align}
Here, $\theta_g$ are the parameters of the generator network. While training the generator, we essentially want to maximize the probability of samples it produces to match the data. Hence we want to maximise $p_{data}(G(z;\theta_g))$.

A discriminator network $D$ on the other hand simply take a data sample $x$ as input and outputs the probability $D(x,\theta_d)$ of the sample coming from the distribution $p_{data}$ rather than it being generated by the generator. $\theta_d$ is the parameter of the discriminator.

Now, the generator wants to generate samples from data distribution. So it must train its parameters so that the generated samples can fool the discriminator. i.e
\begin{equation}
    \min_{\theta_g}\mathbb{E}_{z\sim p_Z}\left[\log\left(1-D(G(z))\right)\right]
\end{equation}

The discriminator, on the other hand, must learn to tell generated and real samples apart. So it must maximize the probability value assigned to actual data samples and minimize the probability value assigned to generated samples.
\begin{equation}
    \max_{\theta_d}\mathbb{E}_{x\sim p_{data}}\left[\log D(x)\right] + \mathbb{E}_{z\sim p_Z}\left[\log\left(1-D(G(z))\right)\right]
\end{equation}

Both the generator and the discriminator networks are trained alternatively so that they try to fool each other. The whole process converges when generator eventually learns to generate samples from $p_{data}$

We use adversarial training in a slightly modified way. Instead of having a generator which maps from input noise to samples to a data distribution, we have a `generator' that takes a noisy image and `generates' the corresponding clean image. This network is essentially a denoiser.

Now the discriminator network has to discriminate between clean images and denoised images. The adversarial network is trained such as to find optimum parameters satisfying
\begin{align}
    \theta^*_g,\theta^*_d = \min_{\theta_g}&\,\max_{\theta_d}\,l_{adv}
    \label{eqn:advloss}
\end{align}
Where the loss function is given by 
\begin{align}
     l_{adv}&=\log D(I_c) + \log(1-D(G(I_n)))
\end{align}

Here, $I_c$ is the clean image and $I_n$ is the noisy image

Eq.~(\ref{eqn:advloss}) corresponds to using a binary cross entropy loss on the output of the discriminator that is trained to tell whether the input belongs to one of the two class; true samples or generated samples.

\begin{figure}[htbp]
    \centering
    \includegraphics[width=0.45\textwidth]{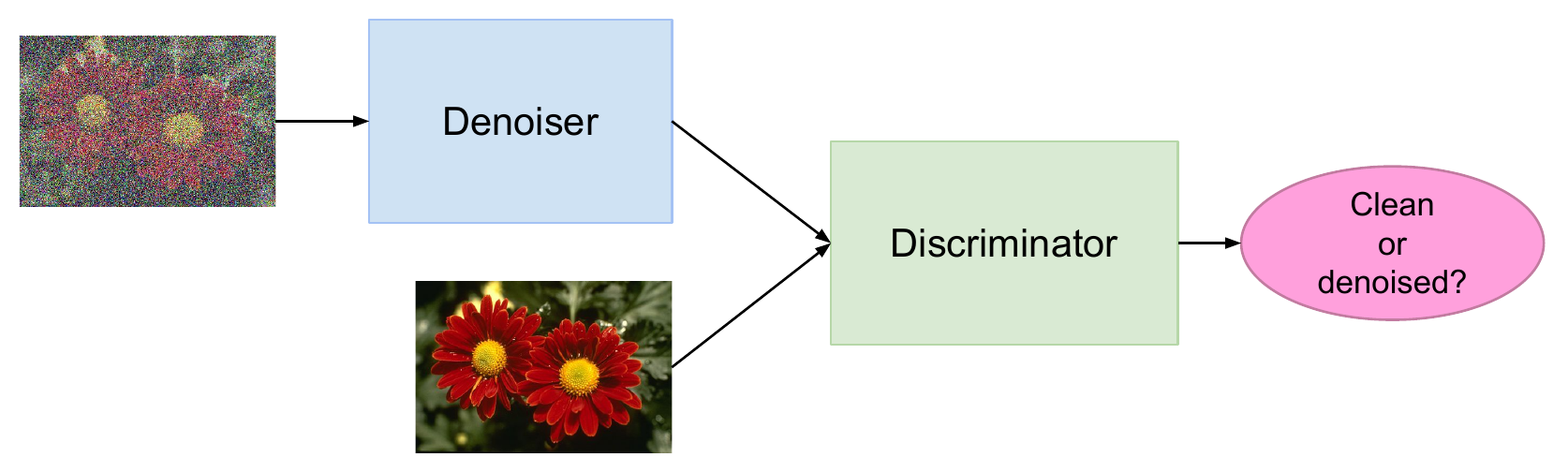}
    \caption{Adversarial training model}
    \label{fig:adversarial}
\end{figure}

But this model allows the generator/denoiser to transform noisy image to any image which the discriminator will classify as a true sample. But for correct denoising, we need the output of denoiser to be very close to the clean image. So we restrict the output of the denoiser to be close to clean image by imposing an extra loss term,
\begin{equation}
    l_{deno} = \frac{\|I_d-I_c\|_2^2}{|I_c|}
\end{equation}
$I_d = G(I_n)$ is the denoised image and $|I_c|$ is the size of the image. This is essentially mean squared loss which penalises  any deviation from the original data (here $I_c$).

Several modifications of adversarial training has been proposed in the literature~\cite{radford2015unsupervised,imporvedGANTraining}, but the idea to use adversarial training for other tasks other than image generation is  not new
~\cite{denton2015deep,mathieu2015deep,ledig2016photo}. But to the best of our knowledge, ours is the first work that uses adversarial training for blind image denoising.

We have used VGG19~\cite{simonyan2014very} model as the discriminator in our experiments. The fully connected layers were replaced by three new layers of size $2048$, $1024$ and $2$ initialized with random weights. 
Then, these layers are fine-tuned to distinguish between the denoiser output and the clean image.

In VGG19 model, the feature detectors (convolutional layers) are kept unmodified throughout the training and only the fully connected layers are allowed to be trained/modified. The discriminator is pre-trained on the denoiser output and the clean image for 10 epochs which gave a cold start accuracy of about $95\%$.

After the noisy-to-clean training phase, the denoiser model can already denoise images to some extent. Since adversarial training is very sensitive to the balance of ability of generator and discriminator, the loss function is modified to accommodate this. 
Essentially, the loss function is the weighted sum of $l_{adv}$ and $l_{deno}$ as follows.

\begin{align}
    loss = l_{deno} + \left(\frac{1+st}{T}\right)l_{adv}
    \label{eqn:loss}
\end{align}

Where $s=0.99$ is a damping factor , $t$ is the iteration number and $T$ is total number of iterations. This ensures that the adversarial loss is weighted less in the beginning of this phase, but as the training progresses, its contribution to loss increases. This weighing scheme allows the discriminator to slowly learn the difference between denoised image and clean image in the initial iterations. Without this weighing scheme, we have observed that the denoiser model quickly starts to generate images to confuse the discriminator rather than trying to produce noise free images. Essentially, it allows the denoiser to strictly stick to denoising rather than trying prematurely to fool the discriminator.

\begin{algorithm}
\caption{Steps for training the adversarial network. $X$ is a set of clean images in the dataset}\label{alg:adv}
\begin{algorithmic}[1]
\Procedure{Adversarial Training}{$X$}
\While{$t<T$}
    \State $x = minibatch(X)$
    \State $\hat{x} = addnoise(x)$
    \State $y = G(\hat{x})$
    \State \parbox[t]{0.38\textwidth}{Train discriminator so that all of $x$ is classified as \emph{true} samples and all of $y$ is classified as \emph{false} samples.}
    \State \parbox[t]{0.38\textwidth}{Train generator/denoiser so that $D(G(\hat{x}))$ always evaluates to \emph{true}.} 
    \State Update loss function according to Eq.~(\ref{eqn:loss})
\EndWhile
\EndProcedure
\end{algorithmic}
\end{algorithm}

We have observed that keeping accuracy of discriminator above $95\%$ helps the model learn faster and hence for ensuring this, in each iteration, the discriminator is shown the data twice. We have used Adam optimizer~\cite{adamoptimizer} for both networks and set the learning rate of the adversarial network to be $10^{-5}$ and the discriminator network to be $10^{-6}$. The procedure for adversarial training is enumerated in Algorithm~\ref{alg:adv}.

\subsection*{Connection of adversarial training to patch prior model}

Adversarial training is motivated by the fact that the final loss function that our adversarial model minimizes is very similar to the loss function derived from patch prior models ~\cite{roth2005fields,schmidt2014shrinkage,zoran2011learning}.

The patch prior model for denoising is given by

\begin{align}
    p(M(I_n)|I_n) = \frac{p(I_n|M(I_n))p(M(I_n))}{Z}
\end{align}
where $M$ is the denoiser model and $M(I_n)$ is the output of the model for a noisy image $I_n$. $Z$ is a normalizing factor.

Assuming Gaussian noise and taking log likelihood, the loss function is given by
\begin{align}
    er[M(I_n),I_n] = \|M(I_n) - I_n\|_2^2 - \frac{1}{C}\log p(M(I_n))
\end{align}
where $C$ is a constant resulting from noise parameters.

In the adversarial model, if we use binary cross-entropy as the loss function for the discriminator and constraint the output of denoiser to be close to the clean image, the model then is optimized over a similar loss function. The only difference being that the output of the network is constrained to be close to the clean image $I_c$  other than $I_n$. This difference is justified as the patch prior models want the output to be close in structure to the actual image, but it doesn't have the clean patch.

\section{Experiments}\label{sec:exp}

In this section, we present the observations made during the training and evaluation of our model for denoising.

\subsection{Training and testing Data}

\textbf{Training Data:}
The training data consists of Images from MIT Indoor dataset~\cite{quattoni2009recognizing} and Places dataset~\cite{zhou2014learning}.
These two datasets were chosen because they contain images of two different modalities; indoor scenes and outdoor scenes. Together, these two datasets have provided our model with good examples of most possible textures and patterns available in real world data.

For preparing training data, we have randomly chosen $5000$ images from each of these datasets. A random $64\times64$ crop is extracted from each of the images. Then the pixels are rescaled to the range $[0,1]$. 

During the training process, the noisy images are generated by adding a random level of Gaussian noise to the image. The model is not given any information about the amount of noise added. This has helped our model to be a blind denoiser.

\textbf{Test Images:}
The model performances are evaluated on the test set used in ~\cite{Vemulapalli_2016_CVPR}. This set of $300$ images contains $100$ images from BSDS300~\cite{MartinFTM01} and $200$ images from PascalVOC~\cite{pascal-voc-2011}. These set of images are a super-set of the test set used in ~\cite{roth2005fields,schmidt2014shrinkage,zoran2011learning} and was first used in \cite{Vemulapalli_2016_CVPR}. Since the denoiser network is fully convolutional, images need not be re-sized or cropped during testing. They can simply be fed to the model and it will reconstruct the denoised image.

\textbf{Validation set:}
We have used the $7$ standard images used in ~\cite{portilla2003image} as the validation set during training procedures. 
During training, the model is evaluated for denoising using each of these images for multiple noise levels. All the denoising performances of the model during training has been plotted by the average performance over these images.

\subsection{Denoising performance}\label{sec:results}

\begin{figure}
    \centering
    \includegraphics[width=0.5\textwidth]{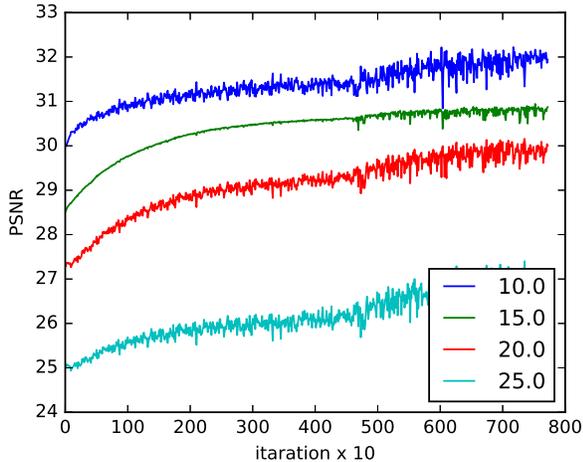}
    \caption{Denoising performance of the denoiser model on validation set during adversarial training. The model performance is evaluated every 10 iteration on each noise level on all $7$ images of the validation set. The plotted values are average of all $7$ PSNR's}
    \label{fig:advValidation}
\end{figure}

\begin{figure}
    \centering
    \includegraphics[width=0.5\textwidth]{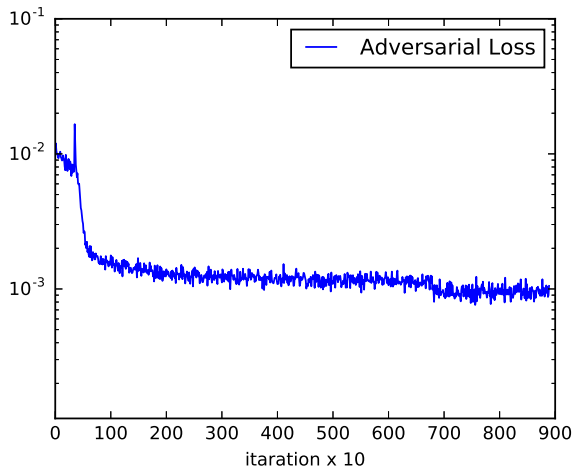}
    \caption{Trend of evolution of Adversarial loss during training iterations. The values are plotted every $10$ iterations.}
    \label{fig:advTrainingError}
\end{figure}

Peak Signal-to-Noise Ratio (PSNR) is a common measure to gauge denoising performance.
PSNR measures dissimilarity between two images and hence to measure denoising performance, we simply measure the PSNR value between the denoised image and the original, noise free image.
For a clean image $I_c$ and a denoised image $I_d$ with range of pixel values from $0$ to $255$, PSNR is computed as
\begin{align}
    PSNR(I_c,I_d) &= 10\log_{10}\left(\frac{255^2}{mse(I_c,I_d)}\right)\\
    mse(I_c,I_d) &= \frac{1}{|I_c|}\|I_c-I_d\|^2_2  \\
    |I_c|&\to\text{ size of the image}
\end{align}

During the initial phase of adversarial training, the discriminator accuracy is comparatively lower because the discriminator cannot classify real and denoised images. But as training progresses, the discriminator gets better at this task. The generator (denoiser) now under the influence of adversarial loss, slowly begins to produce natural looking images and we see a decrease in the training loss. The adversarial loss value vs iteration number is plotted in Fig~\ref{fig:advTrainingError}. 
Fig~\ref{fig:advValidation} shows the average PSNR values over the validation set for each of the noise levels. 

Table~\ref{tab:comparison} gives the performance of our model against other denoising algorithms. A point to be noted here is that except ~\cite{Vemulapalli_2016_CVPR} and our method, all the other methods are not blind denoising techniques. They are provided standard deviation of the added Gaussian noise and the algorithm adapts to these values accordingly.

\begin{figure*}[thbp]
\begin{center}
    \includegraphics[width=\textwidth]{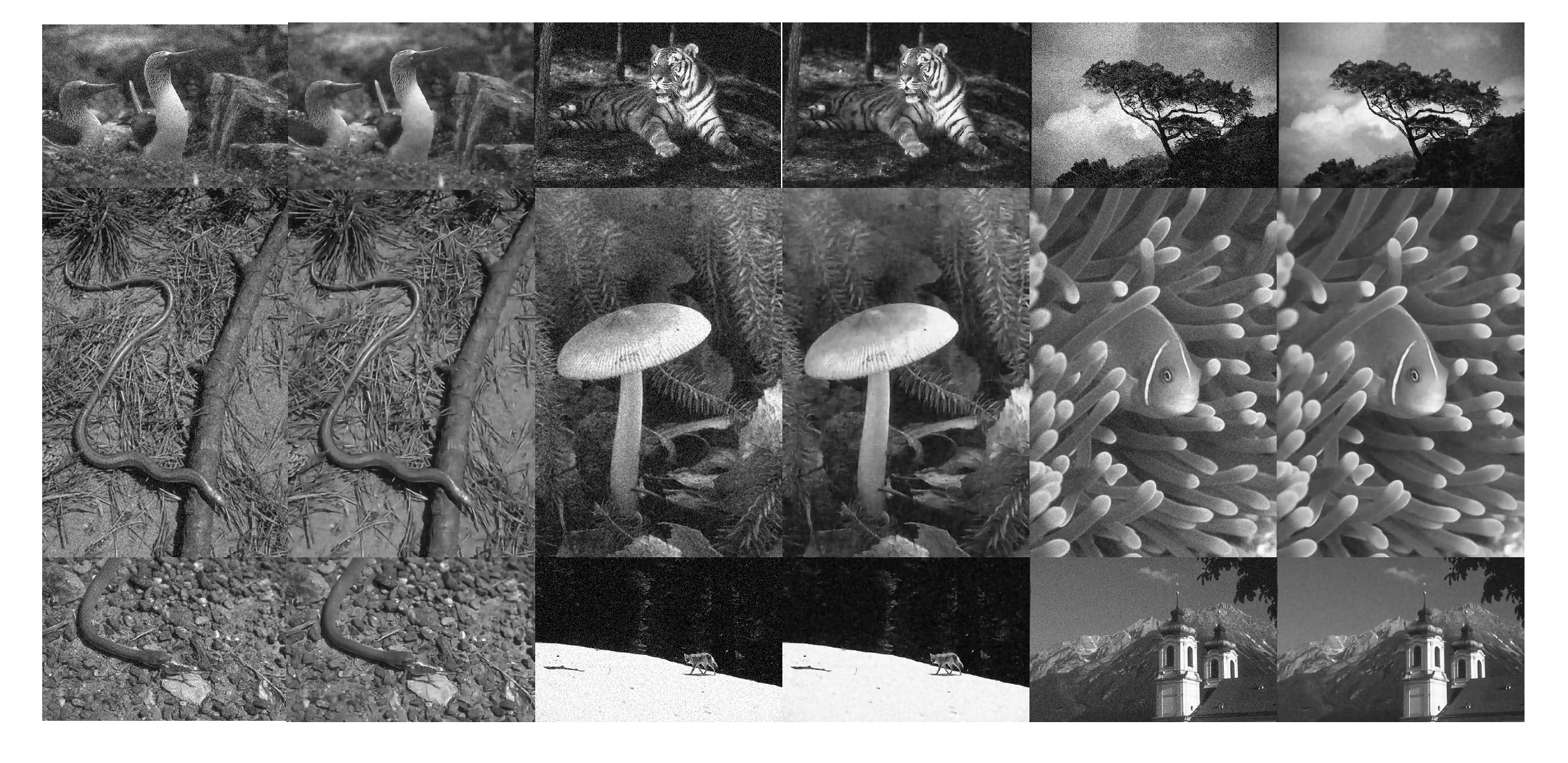}
\caption{Denoising results of our model. Image in the left of each pair shows the noisy image and the image in the right shows the denoised image.}
\label{default}
\end{center}
\end{figure*}

\begin{table}[ht]

\caption{PSNR values of denoised images on test set introduced by~\cite{Vemulapalli_2016_CVPR}. Only DCGRFN\cite{Vemulapalli_2016_CVPR} and our method are blind denoising approaches. Other methods are explicitly given standard deviation of the additive gaussian noise.}
\begin{center}
\begin{tabular}{lllll}
\hline
Sigma                             & 10    & 15    & 20    & 25    \\\hline
BM3D~\cite{dabov2009bm3d}          & 33.38 & 31.09 & 29.53 & 28.36 \\
WNNM~\cite{gu2014weighted}         & 33.57 & 31.28 & 29.7  & 28.50 \\
EPLL~\cite{zoran2011learning}      & 33.32 & 31.06 & 29.52 & 28.34 \\
CSF~\cite{schmidt2014shrinkage}    &  -   &  -    &   -   & 28.43 \\
DCGRFN~\cite{Vemulapalli_2016_CVPR}& 33.56 & 31.35 & 29.84 & 28.67 \\\hline
Ours                              & 33.41 & 31.17 & 29.59 & 28.49 \\
\hline
\end{tabular}
\end{center}
\label{tab:comparison}
\end{table}


\section{Related Work}\label{sec:related}

The corrupting process that results in a noisy image can be seen as
\begin{align}
    I_n = I_c+N
\end{align}
where $N$ is the noise and $I_c$ is the clean image(patch).

If the corrupting noise is uncorrelated, and we have a large number of corrupted samples of the same patch, averaging them all, would give us a very good approximation to the clean patch. But a naive application of this idea is limited by two constraints.
\begin{enumerate}
    \item Large number corrupted versions of same patches are not available. 
    \item We are limited to working with only noisy patches. 
\end{enumerate}
But natural images are full of repeating patterns and textures. The second constraint limits identifying the patterns because high similarity might as well be induced by noise or vice-versa. Solutions to solve these problems have given some of the classical works in denoising.

If we ignore the fact that similarity measure might give incorrect results for noisy patches, then the averaging step has to compensate. A simple Euclidean distance in the local neighborhood will give a set of noisy patches that are similar to each other.

Non-local means algorithm~\cite{buades2005review} modifies the averaging step to be a weighted averaging, where the weights are given by the similarity measure. BM3D~\cite{dabov2009bm3d} uses collaborative filtering of all the similar patches to achieve superior results. Weighted Nuclear Norm Minimization~\cite{gu2014weighted} exploits the fact that set of similar patches would be of low rank if they were noise free. Simply solving for a set which gives a lower weighted nuclear norm removes the noise from the data.

Assuming prior on image patches has lead to denoising methods which does not involve finding similar patches at all. K-SVD~\cite{elad2006image} method applies a sparse dictionary model to noisy patches which essentially remove the noise from them. The sparse dictionary used in this method was `learned' out of the large corpus of natural or clean images.

The first attempt to learn a generic image prior was given by Product-of-Experts~\cite{hinton1999products} which was later extended to image denoising and inpainting by Field-of-Experts~\cite{roth2005fields}. Both methods involve learning a prior from a generic image database and then using the prior for iterating towards a noise free patch. Minimizing the expected Patch Log Likelihood~\cite{zoran2011learning} also used a learned Gaussian mixture prior.

But with deep learning techniques, new methods are devised which can learn image prior implicitly as model parameters and simply compute the noise free patch.  A network resembling fully convolutional network was used in~\cite{NIPS2008_3506} to get a denoiser model. In~\cite{burger2012image}, a 5 layer fully connected network gave state-of-the-art performance. But both these models require different parameters to be specifically trained for each noise level.

In~\cite{Vemulapalli_2016_CVPR}, the authors have used an end-to-end trainable network which uses Gaussian conditional random field. This model uses successive steps of denoising and noise parameter estimation to eventually give a model which can do blind denoising.

In contrast to the existing works, our model is simple and easy to train. 
It essentially results in a set of convolution and non-linearity and hence using it for denoising is extremely simple. Also, our model is not applied on patches. It takes as input the entire image and simply computes the denoised image. This allows it to be fast in comparison. 
The model is trained on varying noise levels together and hence it allows our model to be a  blind denoiser which is trained end-to-end. There is no parameter estimation and the model is capable of automatically adjusting to the required noise level to give the best output.

    
\section{Conclusion}\label{sec:conclusion}

In this work, we addressed whether Convolutional Neural Networks can solve the problem of image denoising.

We have proposed a simple architecture which gives very competitive denoising results.
The architecture contains three unique parts. A multi-scale feature extraction layers, damping layers, and reconstruction layers. 

We have also proposed a three stage training procedure to train the model.
In the first stage, the multi-scale feature extraction layer is trained to extract features for image reconstruction by using clean images.
In the second stage, the damping layers are trained to diminish activations of noise variant filters.

In the final stage, we have successfully adopted adversarial training to this framework with a modified adversarial loss which greatly improves the performance of the denoiser over the limit imposed by $\ell_2$ loss.
The proposed denoiser, a fully convolutional neural network, is a simple model with fewer parameters. 
The model denoises the given noisy image in a single pass without any need for patch extraction step and hence is computationally very efficient.

\section{Acknowledgement}
This work was supported by ISRO-IISc Space Technology Cell, Indian Institute of Science, Bangalore (Project No: ISTC0338).
{
\bibliographystyle{ieee}
\bibliography{bibfile}
}


\end{document}